\documentclass[runningheads]{llncs}

\usepackage{eccv}

\usepackage{eccvabbrv}

\usepackage{graphicx}
\usepackage{booktabs}
\usepackage{makecell}

\usepackage[accsupp]{axessibility}  

\usepackage{hyperref}

\usepackage{orcidlink}

\newcommand{\shortname}{\text{LLaVA-score}}
\begin{document}

\title{Removing Distributional Discrepancies in Captions Improves Image-Text Alignment}

\titlerunning{LLaVA-score}

\author{
Yuheng Li\inst{1}\orcidlink{0000-0001-6082-6398} \and
Haotian Liu\inst{1} \and
Mu Cai\inst{1}\orcidlink{0009-0008-7967-9752} \and
Yijun Li\inst{2}\orcidlink{0000-0001-7295-8750} \and
Eli Shechtman\inst{2}\orcidlink{0000-0002-6783-1795} \and
Zhe Lin\inst{2}\orcidlink{0000-0003-1154-9907} \and
Yong Jae Lee\inst{1}\orcidlink{0000-0001-9863-1270} \and
Krishna Kumar Singh\inst{2}\orcidlink{0000−0002−8066−6835}
}

\authorrunning{Y.~Li et al.}

\institute{University of Wisconsin-Madison \and
Adobe Research 
}

\maketitle

\begin{abstract}

In this paper, we introduce a model designed to improve the prediction of image-text alignment, targeting the challenge of compositional understanding in current visual-language models. Our approach focuses on generating high-quality training datasets for the alignment task by producing mixed-type negative captions derived from positive ones. Critically, we address the distribution imbalance between positive and negative captions to ensure that the alignment model does not depend solely on textual information but also considers the associated images for predicting alignment accurately. By creating this enhanced training data, we fine-tune an existing leading visual-language model to boost its capability in understanding alignment. Our model significantly outperforms current top-performing methods across various datasets. We also demonstrate the applicability of our model by ranking the images generated by text-to-image models based on text alignment. Project page: \url{https://yuheng-li.github.io/LLaVA-score/}

\end{abstract}

\section{Introduction}
\label{sec:intro}

Recent years have seen rapid advances in multimodal research, encompassing both visual generation~\cite{DALLE2, rombach2022high, openai2023DALLE3} and visual understanding~\cite{liu2023llava, alayrac2022flamingo, openai2023GPT4V}. Training capable multimodal models generally requires extensive datasets of image-text pairs, e.g., LAION~\cite{schuhmann2022laion}, CC12M~\cite{changpinyo2021cc12m}, and others~\cite{gadre2024datacomp, sharma-etal-2018-conceptual}, which are collected on a web-scale and thus tend to be noisy. This noise in the data contributes to certain challenges. For instance, vision-language models often struggle with hallucination~\cite{li2023evaluating} and face difficulties in mastering compositional reasoning~\cite{thrush2022winoground,yuksekgonul2023when}. Similarly, text-to-image models frequently fail to generate accurate images when processing complex sentence prompts~\cite{feng2023trainingfree}.

Given these challenges, the ability to automatically assess whether an image and a caption are semantically aligned plays a crucial role. This capability is essential not only for cleaning the data used to pre-train these models but also for evaluating and enhancing the performance of both text-to-image and image-to-text generation models. The most frequently employed metric for this task is the CLIP score~\cite{hessel-etal-2021-clipscore}, which calculates the cosine distance between the CLIP~\cite{radford2021learning} embeddings of the paired text and image. However, it has been observed that CLIP, along with other methods such as BLIP~\cite{li2022blip} and Flava~\cite{singh2022flava}, tends to operate on a bag-of-words basis. They sometimes even cannot tell the difference between simple cases, such as \textit{horse eating grass} versus \textit{grass eating horse}~\cite{yuksekgonul2023when, lin2023visualgptscore, hsieh2023sugarcrepe}.

\begin{figure}[t!]
    \centering
    \includegraphics[width=0.95\textwidth]{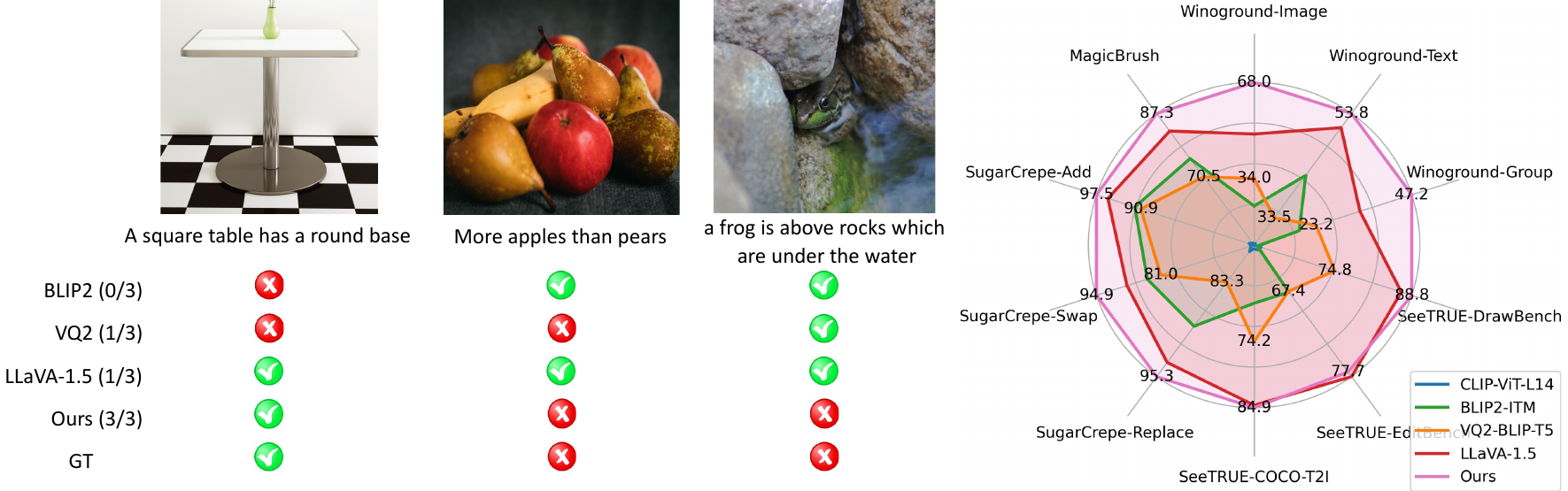}
    \vspace{-0.1in}
    \caption{
\textbf{Left:} Qualitative examples for image-text alignment prediction, where our approach can distinguish fundamental concepts such as positioning, counting, and attributes.
\textbf{Right:} Our approach shows superior performance in image-text alignment.
}
    \label{fig:radar}
    \vspace{-0.1in}
\end{figure}

The conventional approach to training image-text alignment models involves generating negative captions that are misaligned with the images, which are paired with the original positive captions as the training data. For example, prior works randomly shuffle words to create negative captions~\cite{yuksekgonul2023when}, or employ language models to generate coherent negative sentences~\cite{seetrue}.  We follow the latter and leverage an LLM to generate a mix of different types of negative captions during training. Specifically, this includes the \textit{replacing} type, where one linguistic element is substituted with an arbitrary counterpart (\eg `\textit{a \underline{knife} is on the table}' $\xrightarrow{}$ `\textit{a \underline{spoon} is on the table}'), and the \textit{swapping} type, where words within the same sentence are rearranged (\eg `\textit{an \underline{apple} is to the left of a \underline{banana}}' $\xrightarrow{}$ `\textit{a \underline{banana} is to the left of an \underline{apple}'}). The former type can aid in enhancing an image-text alignment model's perceptual skills as it needs to distinguish between the original and replaced elements, while the latter type can help the model's reasoning capabilities as elements in the caption remain the same but the relationship between them changes.

However, critically, we find that this standard approach of generating negative captions, and ensuring that they have coherence (e.g., proper grammar) on a \emph{per-instance} level, is insufficient.  In particular, this approach cannot ensure consistency at the \emph{distribution} level between positive and negative captions.  Distributional biases may originate from the initial dataset or from the rules or models employed to generate negative captions. For instance, the COCO dataset~\cite{le2024coco} contains significantly more captions including the word \textit{giraffe} compared to \textit{elephant}. Yet, when using GPT to generate negative captions, we observe a tendency for GPT to substitute \textit{giraffe} with \textit{elephant}, resulting in a surplus of \textit{giraffe} mentions in positive captions but more \textit{elephant} in negative captions. Unfortunately, this means that an image-text alignment model trained on such data can be biased to predict sentences containing \textit{elephant} to be a positive caption and those that contain \textit{giraffe} to be a negative caption, independent of the paired image.  To address this imbalance, our approach consists of fine-tuning a classifier on text captions only (without paired image inputs) to remove biased data that the text classifier can correctly predict with high confidence. It's important to note that this bias is not unique to the GPT model or how one prompts it, but can arise in other rule-based methods or pre-trained models as well. 

While previous research~\cite{hsieh2023sugarcrepe} noticed distribution differences due to implausible and non-fluent negative captions, our work is the first to eliminate the dataset-level distribution differences using a language model to implicitly encompass all linguistic aspects, without the need for explicit identification. This approach allows us to incorporate considerations like word frequency which is not identified beforehand, thereby providing a more holistic optimization over the dataset's distribution.

For our image-text alignment model, we leverage a state-of-the-art vision-language model, LLaVA~\cite{liu2023llava, liu2023improvedllava}, and finetune it with our curated data. Our model achieves significantly better results compared with other image-text alignment models, demonstrated in Figure~\ref{fig:radar}. We also demonstrate that our curated data can improve other models like BLIP2~\cite{Li2023BLIP2BL}.  Finally, we further show that our image-text alignment model can help with other vision-language tasks like ranking generated images from T2I models~\cite{hu2023tifa}. In summary, we have four main contributions:

\begin{itemize}
\item We identify new dataset-level distribution differences between positive and negative captions that lead to biased image-text alignment models. 
\item To address this, we propose a method to maintain consistency between positive and negative caption distributions, which is critical to ensure that an image-text alignment model relies on both image and text to measure alignment instead of only the biases present in the text.
\item We use our curated training data to finetune existing visual-language models like LLaVA, and obtain state-of-the-art results for image-text alignment. 
\item In addition, we demonstrate the application of our image-text alignment model in ranking the image generations produced by generative models.
\end{itemize}

\section{Related Work}
\label{sec:related}

\textbf{Challenges in Compositional Understanding.}
Image-text pairs form a crucial interface between visual and linguistic modalities, thus evaluating if a given image-text pair is aligned is important for both data curation and model performance evaluation. The pioneering work of CLIP~\cite{radford2021learning} demonstrated this by leveraging an extensive corpus of such pairs for image-text contrastive training. More recent works such as BLIP~\cite{li2022blip, Li2023BLIP2BL} and LLaVA~\cite{liu2023llava, liu2023improvedllava} further utilize large language models (LLMs) to achieve image-text alignment via the text generation objective. Such models usually inherit the frozen CLIP visual encoder to produce a set of visual tokens, and then feed such tokens and the language instructions into the LLM.

It has been observed that vision-language models like CLIP have limited capability in understanding compositionality~\cite{yuksekgonul2023when, thrush2022winoground, zhao2022vl, ma2022crepe, ray2023cola}. Specifically, they find it challenging to recognize the permutation of words within sentences~\cite{thrush2022winoground}. Moreover, these models often struggle to identify the binding of attributes to multiple objects within a single sentence or to discern the relationships between objects~\cite{zhao2022vl, yuksekgonul2023when, ray2023cola}. While models built on top of large language models (LLMs), such as BLIP~\cite{li2022blip} and LLaVA~\cite{liu2023llava}, exhibit improved understanding capabilities, they still face difficulties with complex compositional understanding largely due to their lack of training on sufficiently challenging data.

\textbf{Enhancing Vision-Language Compositionality.}
Several studies focus on improving models' understanding of language compositionality~\cite{yuksekgonul2023when, lin2023visualgptscore, seetrue, le2024coco,doveh2024dense,doveh2023teaching,herzig2023incorporating,zhang2024countercurate}. Some strategies~\cite{seetrue, hu2023tifa} involve using language models to decompose text into multiple succinct assertion phrases, which are then evaluated by VQA models on an individual basis. \cite{lin2023visualgptscore} also finds that assessing the conditional probability of predicting text based on the input image offers more accurate outcomes than traditional discriminative approaches like contrastive or classification scores used in BLIP~\cite{li2022blip}. Still, the most widely used method remains the explicit fine-tuning of models to differentiate between hard negatives and correct  captions~\cite{yuksekgonul2023when,doveh2023teaching, seetrue}. \cite{yuksekgonul2023when} randomly shuffles words to generate negative captions, but subsequent analysis by \cite{hsieh2023sugarcrepe} points out the approach's flaw: models could simply rely on textual cues (e.g., grammar correctness) for predictions. \cite{seetrue} attempts to employ LLMs to create negative captions, ensuring the negative captions are both fluent and meaningful. Nonetheless, we observed that  current fine-tuning methods do not generate a diverse range of negative prompts. Moreover, merely evaluating the grammar or logical coherence of negative captions is insufficient to eliminate data bias in the distribution of negative captions.

\section{Approach}
\label{sec:approach}

\begin{figure}[t!]
    \centering
    \includegraphics[width=0.9\textwidth]{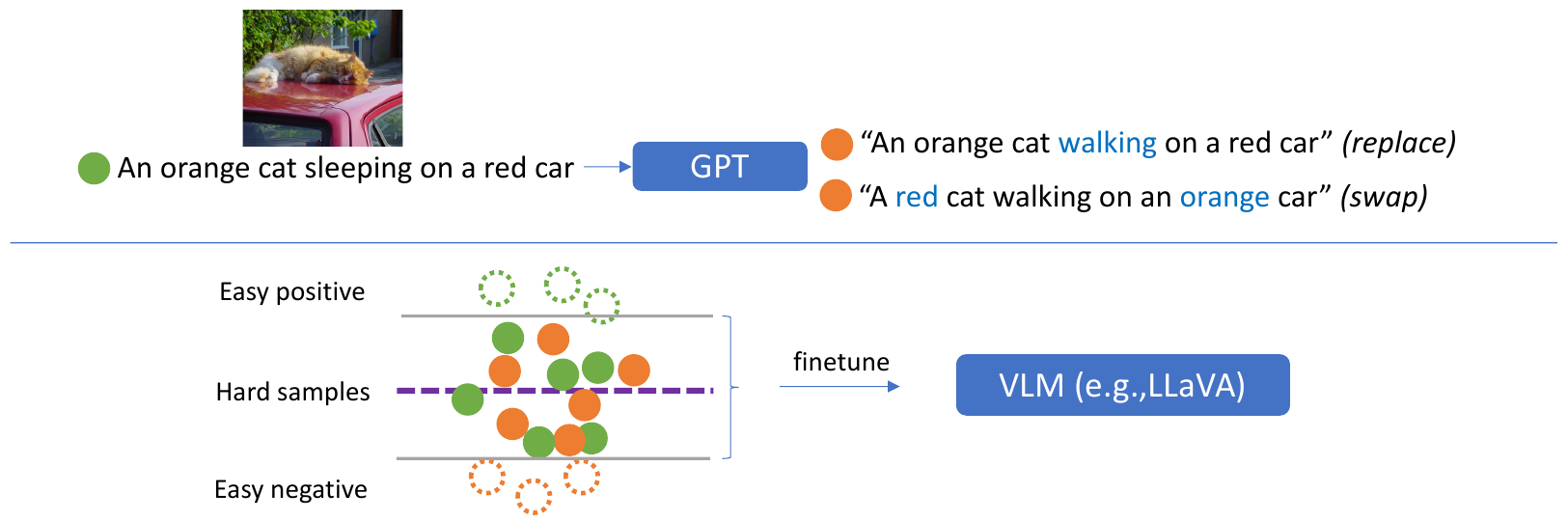}
    \caption{\textbf{Top:} We feed the positive caption (green dots) into GPT to create two types of negative captions (red dots): substituting one linguistic element with any plausible alternative or swapping the positions of two components. The blue part in negative captions highlights the modifications. \textbf{Bottom:} we remove easy negative samples using only text data and utilize the remaining samples to fine-tune vision-language models.}
    \label{fig:approach}
\end{figure}

In our approach, we assume access to images accompanied by accurately labeled positive captions, similar to those found in the MSCOCO~\cite{lin2014microsoft} dataset. Following~\cite{hsieh2023sugarcrepe}, contrary to adopting rule-based techniques that often generate illogical sentences, our method utilizes large language models (LLMs) like GPT4 to transform positive captions into negative ones. We first outline our strategy for generating various types of negative captions and then present a straightforward technique to mitigate biases inherent in the distributions of positive and negative captions. We show the entire pipeline in Figure~\ref{fig:approach}.

\subsection{Constructing Diverse Negative Captions}

Suppose we have a dataset consisting of image-text pairs $\{I, T_p\}$, where $I$ represents an image and $T_p$ is its associated positive caption. Previous research~\cite{yuksekgonul2023when, seetrue} showed that merely randomly shuffling image-text pairs to generate negative samples is insufficient to learn capable vision-language models that properly understand the structure of a caption and its relationship with the image. These studies have underscored the value of constructing hard negative captions to significantly improve the model's language compositional abilities. Following this insight~\cite{seetrue}, we utilize large language models (LLMs) to generate such hard negative captions. Specifically, we create two types of hard negative captions.

\begin{figure}[t!]
    \centering
    \includegraphics[width=0.9\textwidth]{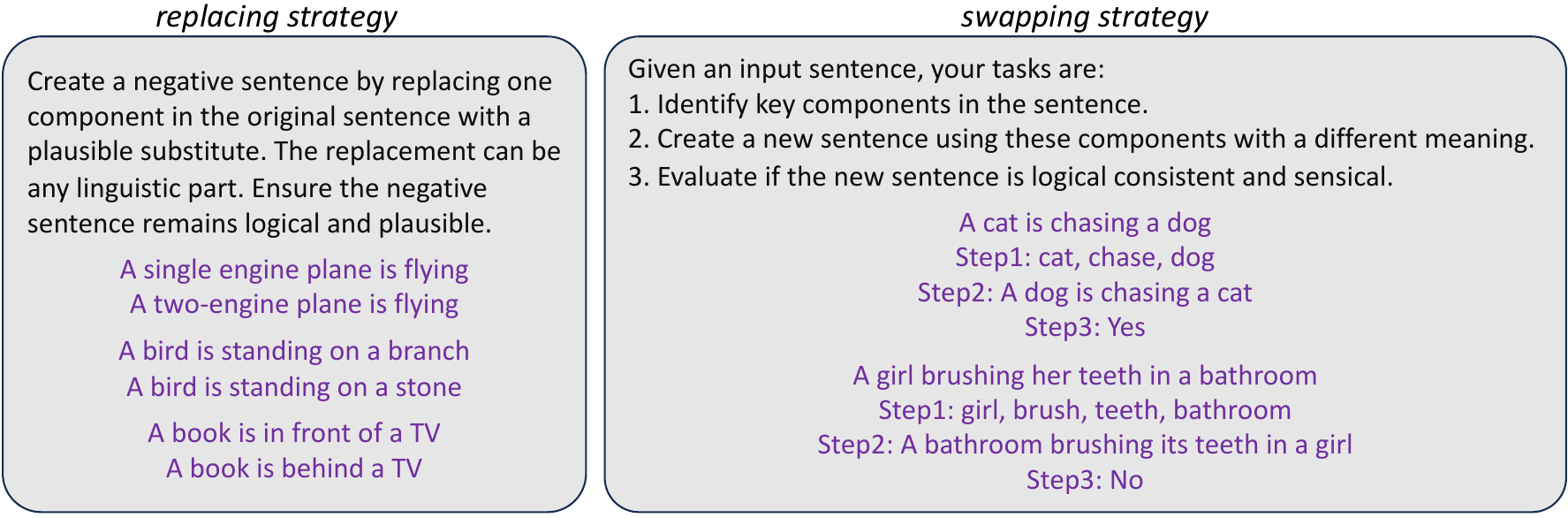}
    \vspace{-5pt}
    \caption{Prompts used in GPT for generating two types of negative captions, with in-context examples shown in purple. 
  }

    \label{fig:negative}
\end{figure}

%
We refer to the first method for creating negative captions as the \textit{replacing} strategy, which identifies key components in a language and uses a language model to replace it with other plausible substitutes. The replaced component can be any linguistic part such as a noun, adjective, preposition, etc. For example, \textit{``a photo of a broken down stop sign''} could be replaced with \textit{``a photo of a brand new stop sign''}; \textit{``a cute cat looking at a bird''} could be changed to \textit{``a cute dog looking at a bird''}. Conceptually, this type of negative caption aims to enhance the model's recognition capabilities. For executing this task, we engage GPT by providing it with specific instructions followed by some contextual examples, as illustrated in Figure~\ref{fig:negative}.

We refer to the second method for creating negative captions as the \textit{swapping} strategy. This approach involves generating a new sentence by utilizing the original language components, typically resulting in the swapping of two or several identical linguistic elements. Specifically, we first ask GPT to break down the original positive sentence into its key components, and then request the model to construct a new sentence with these elements. For example, if the input caption is \textit{``an airplane is flying in the blue sky''}, GPT is tasked with identifying the key components including \textit{``airplane''}, \textit{``flying''}, \textit{``blue''}, and \textit{``sky''}. The newly crafted sentence could be \textit{``a blue airplane is flying in the sky''}. Please note that for some short captions, there may not be enough language elements to form a reasonably different sentence in meaning. Thus, we also need to judge if the new sentence makes sense or not. Please refer to the Figure~\ref{fig:negative} for our prompt used in GPT.

\begin{figure}[t!]
    \centering
    \vspace{-8pt}
    \includegraphics[width=0.9\textwidth]{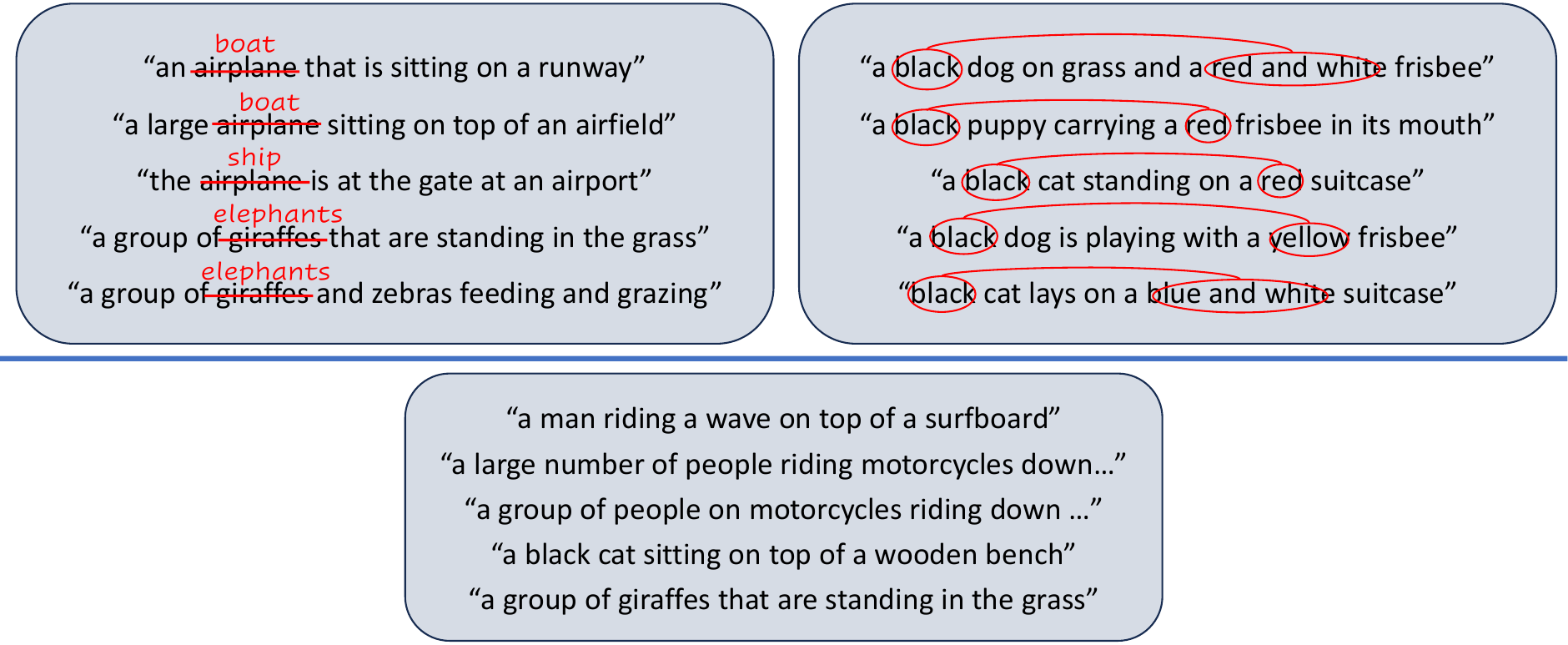}
    \vspace{-5pt}
    \caption{Top prediction based on a text-only binary classifier. \textbf{Top left:} Negative captions generated through replacement strategy. \textbf{Top right:} Negative captions generated through swapping strategy. \textbf{Bottom:} Positive captions from the COCO dataset. }
    \label{fig:text_along_pred}
\end{figure}

\subsection{Addressing Distribution Discrepancies}

Previous studies~\cite{hsieh2023sugarcrepe} have demonstrated that artifact issues arising from incorrect construction in negative captions can lead to distribution disparities between positive and negative captions. To mitigate this, commonsense and grammar models are employed. Nonetheless, our findings indicate that \emph{merely assessing the sensibility of sentences} is insufficient for aligning the distributions of positive and negative sentences because of two following two reasons.

First, each image-pair dataset inherently establishes its own unique distribution. For instance, the COCO dataset~\cite{lin2014microsoft}, despite its widespread use and diverse content, predominantly encompasses everyday scenes, objects, and activities. Common subjects such as people and motorcycles, along with frequent activities like skiing and surfing, characterize the distinct distribution of COCO. Second, when employing rule-based methods or prompting language models to generate negative captions, any strategy adopted will inevitably reflect biases from human preconceptions or pre-trained models. This leads to negative captions originating from a different distribution.

To illustrate this concept, we develop a blind text-only binary classifier that processes positive and negative captions without seeing any images. Surprisingly, we find that despite the coherence and logical structure of negative captions, the classifier is adept at distinguishing between negative and positive captions based solely on the text. Figure \ref{fig:text_along_pred} shows the most confidently correct predictions for both our \textit{replace} and \textit{swap} data categories constructed from COCO. In the dataset with replaced negative data, it is observed that GPT frequently substitutes the terms \textit{"boat"} and \textit{"elephant"} for the original COCO captions' \textit{"airplane"} and \textit{"giraffe"}, respectively. This pattern makes it straightforward for a text-only model to identify these captions as negative. In our \textit{swap} data, the text-only classifier is able to identify captions with colorful animals interacting with a black object as negative. Note that these negative prompts are logically coherent and grammatically correct, thus cannot be easily detected by a grammar model. Conversely, the top accurately predicted positive captions often depict common activities like surfing and motorcycle riding, which align closely with the typical content style of COCO. It's important to recognize that data bias is not solely a characteristic of our method for generating data nor the manner in which we prompt GPT. In the experimental section, we demonstrate that such distribution discrepancy is also present in other approaches~\cite{hsieh2023sugarcrepe} that employ GPT for creating negative prompts.

The presence of bias in the data can obstruct a vision-language model's ability to truly understand image structures and learn language compositions, as the model might rely solely on textual cues for making predictions. To address this issue, we propose a straightforward solution aimed at reducing data distribution differences by selectively removing data that is predicted with high confidence by a text-alone model.
 
Figure~\ref{fig:approach} (bottom) depicts our conceptual goal of data filtering: to eliminate straightforward or biased positive and negative samples. This is done to ensure that the remaining text data cannot be distinguished by any text classifier, based solely on the text information. Essentially, our aim is to maximize the entropy of the text information within our dataset. In practice, we organize our dataset into $N$ partitions. For each iteration, one partition is designated as the test set while the remaining $N-1$ partitions serve as the training set. We employ a pretrained Bert model~\cite{devlin2018bert} as our text-only classifier, and train it on the training set. Subsequently, this trained classifier is applied to the designated test set. We then rank the correctly predicted samples by the classifier's confidence level, removing the top $k\%$ of these samples for both positive and negative class predictions. The rest of the data is retained as our refined dataset. This procedure is repeated for each partition, ensuring a comprehensive reduction of bias across the dataset.

Note that while our observation and data filtering approach is applied to address the image-text alignment issue, it could be a more general problem for any multimodal data scenario where bias in one modality might negatively impact model training.

\subsection{Finetuning VLMs for Image-Text Alignment Scoring}
 \label{approach:finetuning}

Once we have the unbiased data, we opt to fine-tune a vision-language model (VLM) to enhance its language compositional understanding capabilities with respect to images. In our study, we mainly select LLaVA-1.5~\cite{liu2023improvedllava} due to its superior performance in image and text understanding. Since LLaVA-1.5 is designed to generate text, to adapt it for use as a image-text score calculator, we employ the following prompt formatting: \textit{``Does this image $I$ match the following caption $T$. Answer Yes or No directly.''} Given that LLaVA relies on a language model to produce subsequent words, we manually extract the logits associated with the responses \textit{Yes} and \textit{No} for the next word. We then define the matching score as:

\begin{equation}
\frac{e^{\mathbf{P}(\textit{Yes}|prompt)}}{e^{\mathbf{P}(\textit{Yes}|prompt)} + e^{\mathbf{P}(\textit{No}|prompt)}}
\label{eq:probability_ratio}
\end{equation}

We discover that this straightforward approach is quite effective and can already outperform many existing state-of-the-art baselines. However, LLaVA-1.5 was not specifically trained for this type of matching problem. Therefore, to enhance its performance, we finetune it with the same prompt formatting introduced above using our curated data. We assign the labels \textit{Yes} and \textit{No} to positive and negative pairs, respectively.

Our curated dataset is not restricted to LLaVA-1.5.  In our experiments, we also finetune a Q-former equipped with the Image-Text Matching (ITM) head in BLIP2~\cite{Li2023BLIP2BL}. The ITM head is essentially a binary classification layer, identical to our dataset's structure. We thus fine-tune the model using the standard cross-entropy loss.

\section{Results}
\label{sec:result}

We evaluate our fine-tuned LLaVA-1.5 model against various baselines across different datasets. Additionally, we conduct ablation studies to evaluate the impact and importance of our dual-strategy approach for generating diverse negative captions and our method for addressing data distribution discrepancies. Since our main model is built upon LLaVA-1.5, we name our score as \shortname{}.

\subsection{Baselines}

We evaluate our model against a range of leading multimodal models:

\noindent\textbf{CLIP-ViT-L/14}, the largest variant of OpenAI CLIP models~\cite{radford2021learning}.
    
\noindent\textbf{BLIP-2}~\cite{Li2023BLIP2BL}, which incorporates both image-text matching and image-text contrastive learning approaches.

\noindent\textbf{NegCLIP}, as introduced in~\cite{yuksekgonul2023when}, fine-tuned on challenging negative captions generated by randomly shuffling words within sentences.
    
\noindent\textbf{VisualGPTScore}~\cite{lin2023revisiting} proposes utilizing the probability of generating specific text given an image (i.e., $P(T|I)$) as an effective metric for calculating image-text alignment scores. For this, we utilize LLaVA-1.5~\cite{liu2023improvedllava}, the state-of-the-art vision-language model, to calculate their proposed VisualGPTScore, employing a reweighting technique as suggested by~\cite{lin2023revisiting}.

    \noindent\textbf{Image-Reward}~\cite{xu2024imagereward}: A reward model, trained on human preferences of image-text pairs produced by Text-to-Image (T2I) models in DiffusionDB~\cite{wangDiffusionDBLargescalePrompt2022}.
    
    \noindent\textbf{VQ2}~\cite{seetrue} first extracts a set of candidate QA pairs from the text, then uses a VQA model to score each pair. For this baseline, we report results using both the PaLI model~\cite{Chen2022PaLIAJ}—directly citing their paper as the model is not publicly available—and BLIP-T5~\cite{Li2023BLIP2BL} which is accessible via their official GitHub repository~\cite{wysiwyr}. Note that while TIFA~\cite{hu2023tifa} similarly utilizes a QA pairs approach, we exclude it from our results due to VQ2 demonstrating superior performance, aiming for simplicity in our comparison.
    
    \noindent\textbf{PaLI} fine-tuned on the SeeTrue dataset~\cite{seetrue}: A version of PaLI specifically fine-tuned for the alignment task, using a curated dataset. Performance metrics are cited directly from the original paper~\cite{seetrue}.
    
    \noindent\textbf{LLaVA-1.5}~\cite{liu2023improvedllava}, Originally a text generation model, we employ a specific prompt as introduced in Sec.~\ref{approach:finetuning}. By using Equation~\ref{eq:probability_ratio}, we convert it into a scoring function. Through empirical testing, this prompt is found to be the most effective, and it is utilized to finetune our model.

\subsection{Datasets and Metrics}

We evaluate on the following datasets and metrics:

    \noindent\textbf{Winoground}~\cite{thrush2022winoground}: This dataset uniquely comprises quartets, each including two images and two texts, necessitating a nuanced interpretation of both linguistic and visual elements for accurate matching. The metrics reported for this dataset encompass an image score, a text score, and a group score.
    
    \noindent\textbf{SeeTRUE}~\cite{seetrue}: A benchmark designed for assessing vision-language models, featuring a test set combining multiple sources. Current sources include Drawbench, EditBench, and COCO-t2i, pairing real texts with synthetic images. Following~\cite{seetrue}, we utilize ROC-AUC as the evaluation metric.
    
    \noindent\textbf{SugarCrepe}~\cite{hsieh2023sugarcrepe}: A recently introduced benchmark focuses on generating creative negative captions using a language model and employs grammar and common-sense models for data cleaning, marking a novel approach in benchmark design.
    
    \noindent\textbf{MagicBrush}~\cite{Zhang2023MagicBrush}: This benchmark facilitates human-driven image editing, constructed using Dall-E 2, and includes captions for both the original and edited images. Featuring quartets similar to Winoground~\cite{thrush2022winoground}, we adopt the same methodology for calculating the group score with modification due to nature of data.  Refer supp for details.

\subsection{Implementation Details}

For our dataset curation, we select COCO, which provides positive image-text pairs.
To diversify our image dataset and enhance our model's robustness to synthetic images, we incorporate a subset of the training images from SeeTRUE~\cite{seetrue}, specifically the \textit{coco\_train\_t2i} set.
To construct negative data, we use GPT to generate two types of negative captions (\textit{swap} and \textit{replace}), and we perform random sampling to maintain an equal amount of positive and negative data. In the process of curating our dataset, we ensure that neither images nor captions included in the training dataset are present within the test dataset.

For our model development, we choose to refine LLaVA-1.5~\cite{liu2023improvedllava} using our curated data. We train the model with a batch size of 64 on 8 NVIDIA A100 GPUs for a single epoch, setting the learning rate at 2e-6. For our data filtering, we use $k$ to be 30\% and $N$ to be 5.  Find more details in the supp.

\subsection{Main results}

Table~\ref{table:main} presents a comparison between our models and various strong baselines across different datasets. It is evident that our model outperforms others in nearly all datasets. Interestingly, utilizing Eq.~\ref{eq:probability_ratio}, LLaVA-1.5 zero-shot performance is the second-best method overall. Particularly for Winoground, a benchmark well-known for its challenges in visual and linguistic reasoning, some models such as CLIP, BLIP2, and Image-Reward perform at or below chance level. Our fine-tuned LLaVA-1.5 substantially enhancing its reasoning capabilities showing the importance of our strategy for training data curation. For VQ2, the VQA models (BLIP or PaLI) underperform our model due to their lack of training on challenging negative examples. The finetuned PaLI model also falls short, attributed to its training dataset's lack of diversity and absence of data filtering. Our model also shows impressive results on synthetic image benchmarks like SeeTRUE and MagicBrush.

\begin{table}[t]
\centering
\caption{We evaluate our model alongside baselines across multiple datasets. The best results are shown in bold. For three subsets of the SeeTRUE dataset, we present the ROC AUC scores following the original paper~\cite{seetrue}. For the SugarCrepe dataset, accuracy is employed as the performance metric. For the MagicBrush~\cite{Zhang2023MagicBrush} dataset, we report the group score. We use LLaVA-1.5 to calculate the VisualGPT score~\cite{lin2023visualgptscore}.  }
\vspace{-5pt}
\resizebox{\columnwidth}{!}{
\begin{tabular}{@{}lccc|ccc|ccc|c@{}}
\toprule
{} & \multicolumn{3}{c|}{Winoground} & \multicolumn{3}{c|}{SeeTRUE} & \multicolumn{3}{c|}{SugarCrepe}  & {MagicBrush} \\

{} & {image} & {text} & {group} & DrawBench & EditBench & COCO-T2I & replace & swap & add  & {} \\
\midrule
Chance Performance   & 25.00 & 25.00  & 16.67  & 50.0  & 50.0  & 50.0  & 50.0  & 50.0  & 50.0  & 33.33 \\
\midrule
CLIP-ViT-L-14~\cite{radford2021learning}        & 10.50 & 28.50  & ~7.75  & 61.4  & 62.1  & 59.2  & 79.4  & 61.4  & 74.8  & 52.89 \\
NegCLIP~\cite{yuksekgonul2023when}              & 11.75 & 30.75 & ~8.25  & 63.2  & 66.0  & 62.8  & 85.3  & 75.3  & 87.2  & 61.12 \\
BLIP2-ITM~\cite{Li2023BLIP2BL}            & 24.25 & 41.75 & 19.00 & 60.8  & 67.5  & 68.0  & 88.9  & 83.9  & 91.8  & 75.32 \\
BLIP2-ITC~\cite{Li2023BLIP2BL}            & 12.00 & 28.50  & ~8.50  & 64.9  & 67.9  & 69.9  & 86.7  & 66.9  & 92.3  & 67.85 \\
Image-Reward~\cite{xu2024imagereward}         & 15.25 & 43.00 & 12.75 & 70.4  & 70.2  & 77.0  & 88.2  & 81.0  & 95.2  & 70.28 \\
VisualGPT~\cite{lin2023visualgptscore}            & 37.00 & 44.25 & 27.50  & 77.0  & 74.2  & 69.1  & 88.2  & 87.1  & 95.5  & 78.31 \\
VQ2 (PaLI)~\cite{seetrue}         & 42.25 & 47.00 & 30.50 & 82.6  & 73.6  & 83.4  & -     & -     & -     & -     \\
VQ2 (BLIP-T5)~\cite{seetrue}       & 34.00 & 33.50 & 23.25 & 74.8  & 67.4  & 74.2  & 83.3  & 81.0  & 90.9  & 70.46 \\
PaLI (ft on SeeTRUE)~\cite{seetrue} & 38.00 & 46.50 & 28.75 & 86.8  & 77.2  & 83.2  & -     & -     & -     & -     \\
LLaVA-1.5~\cite{liu2023improvedllava}           & 49.75 & 51.00 & 34.25 & 86.9  & \textbf{78.3}  & 84.5  & 93.5  & 88.3  & 95.8  & 82.61 \\
\shortname{} (Ours)               & \textbf{68.00} & \textbf{53.75} & \textbf{47.25} & \textbf{88.8}  & 77.7  & \textbf{84.9}  & \textbf{95.3}  & \textbf{94.9}  & \textbf{97.5}  & \textbf{87.28} \\
\bottomrule

\end{tabular}

}
\vspace{-0.1in}

\label{table:main}
\end{table}

\subsection{Performance on Attribute, Counting, Spatial Reasoning}

In the preceding section, we demonstrated that our model surpasses baseline models across various datasets. This section offers an alternative perspective on our model's performance, particularly in areas where both vision-language models and recent text-to-image (T2I) models encounter significant challenges. These challenges include multiple attributes binding, counting objects, and understanding spatial relationships between objects. Addressing these issues is crucial for both visual understanding and evaluating the capabilities of generative models.

\begin{figure}[t!]
    \centering
    \includegraphics[width=0.85\textwidth]{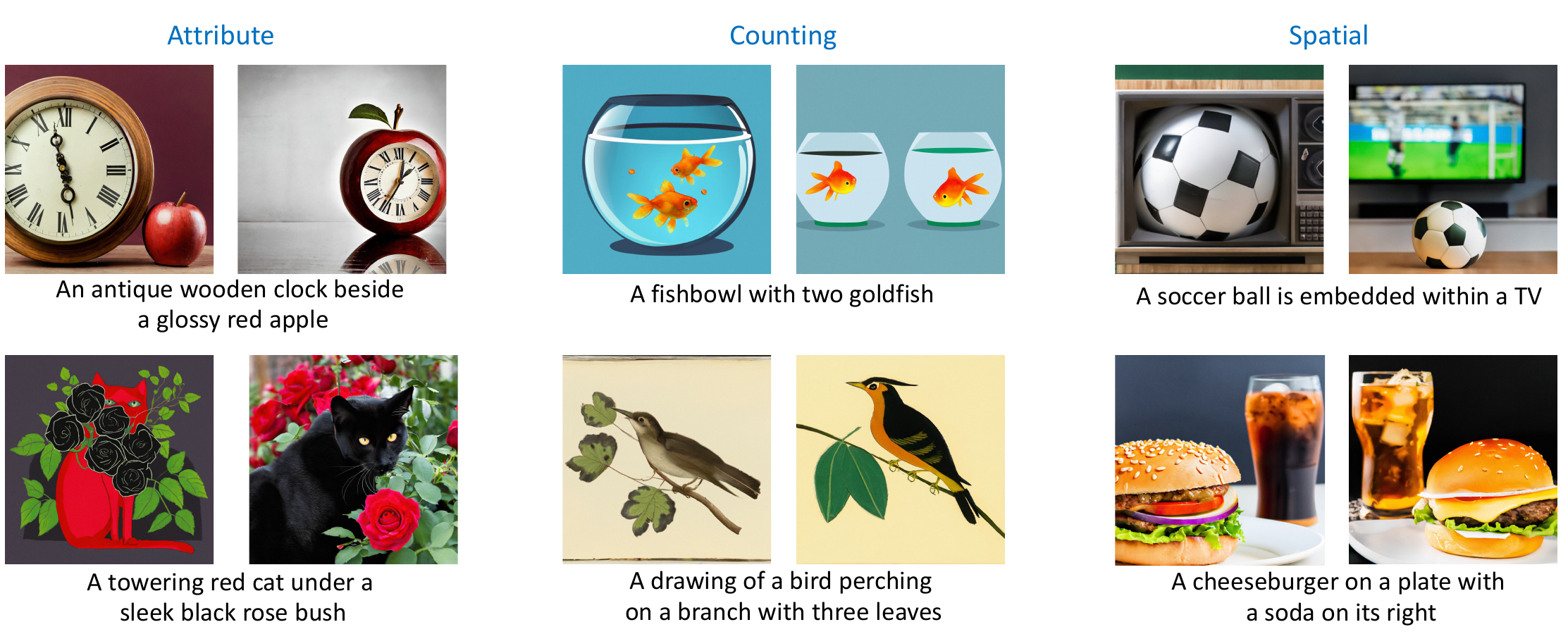}
    \vspace{-0.1in}
    \caption{Our curated test datasets feature captions paired with one positive image and one negative image each. All the positive images are displayed on the left side.}
    \label{fig:mydata}
    \vspace{-0.1in}
\end{figure}

To quantify the improvements our model achieves in addressing these common challenges, we construct three specialized datasets. Specifically, we prompt GPT to generate scenarios involving attribute binding, object counting, and spatial relationships, providing in-context examples to guide the generation process. Subsequently, the T2I model~\cite{if} is employed to generate 50 images for each prompt. From these, we manually select one positive and one negative image based on image-text alignment. We opt for synthetic images as it offers greater flexibility, as illustrated in Figure~\ref{fig:mydata}, it can incorporate diverse styles (e.g., painting styles for birds), create unconventional images (e.g., the examples involving soccer and cats), and has attribute/object merging capabilities of T2I models to enrich our dataset with varied negative cases (e.g., the negative example of mistakenly combining clock and apple).

Table~\ref{table:mydata} presents our results. We report classification accuracy as the metric, alongside showing both positive and negative results in gray color. We categorize the baseline models into two distinct groups for comparison purposes. 1) Upper section of the table, consists of models that generate scores without an established classification threshold. Consequently, we calculate the accuracy by using an oracle threshold, which involves optimizing over all predicted scores to find the best cutoff (cheating on the test set). 2) Lower part of the table, operates with an inherent decision boundary for classification.

Our model demonstrates superior performance across all three challenging cases, beating competitors that perform only marginally better than random guessing, particularly on tasks requiring spatial understanding. It's noteworthy that many models exhibit a bias, often predicting the same outcome across the test dataset. This tendency is underscored by their high accuracy on positive samples contrasted with poor performance on negative ones. Given that our images are synthesized using text-to-image (T2I) models, essential linguistic elements are likely to be present even in images labeled as negative (refer to Figure~\ref{fig:mydata}). This observation suggests that baseline models operate similarly to a bag-of-visual-words approach, predicting an image as matching the textual description as long as it contains certain key visual concepts.

\begin{table}[t!]
\caption{Comparison of our model against baselines across three prevalent challenges, highlighting the highest average accuracy in bold.}
\centering
\scalebox{0.9}{
\setlength{\tabcolsep}{5pt}
\begin{tabular}{@{}lccc|ccc|ccc@{}}
\toprule
{} & \multicolumn{3}{c|}{Attribute} & \multicolumn{3}{c|}{Counting} & \multicolumn{3}{c}{Spatial} \\
{} & avg & pos & neg & avg & pos & neg & avg & pos & neg \\
\midrule
Chance Performance   & 50 & 50 & 50 & 50 & 50 & 50 & 50 & 50 & 50 \\
\midrule
\multicolumn{10}{l}{\it Threshold-Independent Models (with oracle)} \\
CLIP-ViT-L-14~\cite{radford2021learning}           & 63 & {\color{gray} 52} & {\color{gray} 74} & 58 & {\color{gray} 68} & {\color{gray} 48} & 53 & {\color{gray} 18} & {\color{gray} 88} \\
NegCLIP~\cite{yuksekgonul2023when}                 & 65 & {\color{gray} 78} & {\color{gray} 52} & 59 & {\color{gray} 66} & {\color{gray} 52} & 57 & {\color{gray} 48} & {\color{gray} 66} \\
BLIP2-ITC~\cite{Li2023BLIP2BL}               & 66 & {\color{gray} 72} & {\color{gray} 60} & 57 & {\color{gray} 36} & {\color{gray} 78} & 57 & {\color{gray} 90} & {\color{gray} 24} \\
VisualGPT~\cite{Chen2022CVPR}              & 73 & {\color{gray} 90} & {\color{gray} 56} & 65 & {\color{gray} 52} & {\color{gray} 78} & 62 & {\color{gray} 56} & {\color{gray} 68} \\
\midrule
\multicolumn{10}{l}{\it Inherent Decision Models} \\
BLIP2-ITM~\cite{Li2023BLIP2BL}           & 58 & {\color{gray} 100} & {\color{gray} 16} & 53 & {\color{gray} 96} & {\color{gray} 10} & 51 & {\color{gray} 100} & {\color{gray} 2} \\
Image-Reward~\cite{xu2024imagereward}        & 70 & {\color{gray} 100} & {\color{gray} 40} & 61 & {\color{gray} 100} & {\color{gray} 22} & 57 & {\color{gray} 98} & {\color{gray} 16} \\
VQ2 (BLIP-T5)~\cite{seetrue}       & 66 & {\color{gray} 94}  & {\color{gray} 38} & 56 & {\color{gray} 82} & {\color{gray} 30} & 54 & {\color{gray} 94} & {\color{gray} 14} \\
LLaVA-1.5~\cite{liu2023improvedllava}           & 71 & {\color{gray} 98}  & {\color{gray} 44} & 62 & {\color{gray} 96} & {\color{gray} 28} & 57 & {\color{gray} 98} & {\color{gray} 16} \\
\shortname{} (Ours)                & \textbf{81} & {\color{gray} 90}  & {\color{gray} 66} & \textbf{71} & {\color{gray} 86} & {\color{gray} 56} & \textbf{81} & {\color{gray} 76} & {\color{gray} 86} \\
\bottomrule
\end{tabular}
}
\vspace{-0.1in}
\label{table:mydata}
\end{table}

\subsection{Ablation Studies}

\textbf{Importance of different negative data.} Table~\ref{table:ablation} begins with our baseline model, LLaVA-1.5~\cite{liu2023improvedllava}, outlining its performance. The following two rows illustrate the effects of finetuning the LLaVA-1.5 model with only \textit{replace} and \textit{swap} negative data, with filtering processes applied. As shown in the subcategories of the SugarCrepe dataset, \textit{replace} and \textit{swap} data enhance performance in their respective categories but do not significantly impact other cases. Nevertheless, our final model excels across almost all tests, indicating that combining \textit{replace} and \textit{swap} negative data has a complementary effect, particularly noticeable in the \textit{add} category of the SugarCrepe dataset and other evaluated datasets.

\noindent\textbf{Importance of data filtering.} The subsequent row presents outcomes from merging two types of negative data without any filtering. This combination has shown to be beneficial for specific datasets, like Winoground and SugarCrepe, although the results don't match those of the final model with filtering. However, biases in the text data  cause the fine-tuned model to show no improvement on datasets like MagicBrush and SeeTRUE. Since data filtering reduces the amount of data, the fifth row displays results from randomly subsampling a training set to match the quantity of filtered data. This comparison highlights the effectiveness of our filtering technique as without it, the model may get biased to just rely on language distribution and ignore the image when making predictions.

\begin{figure}[t!]
    \centering
    \vspace{-5pt}
    \includegraphics[height=8em]{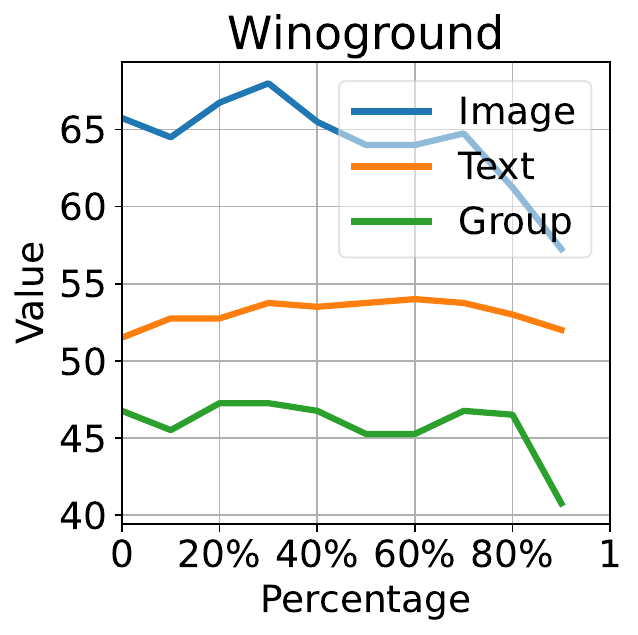}
    \includegraphics[height=8em]{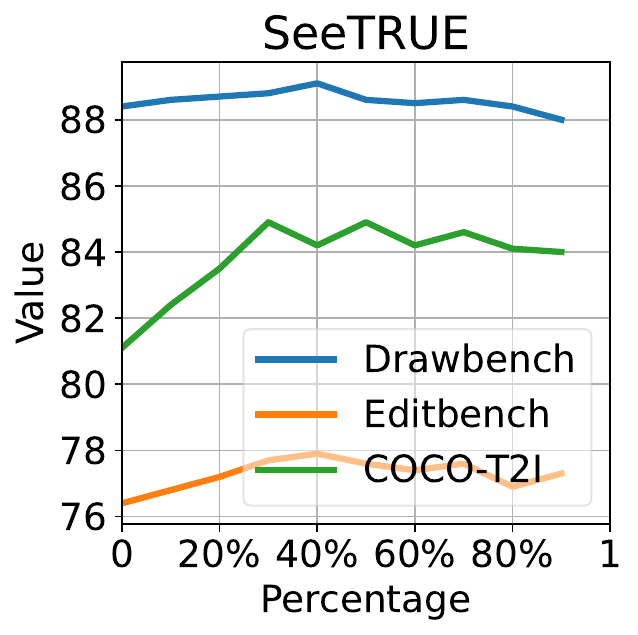}
    \includegraphics[height=8em]{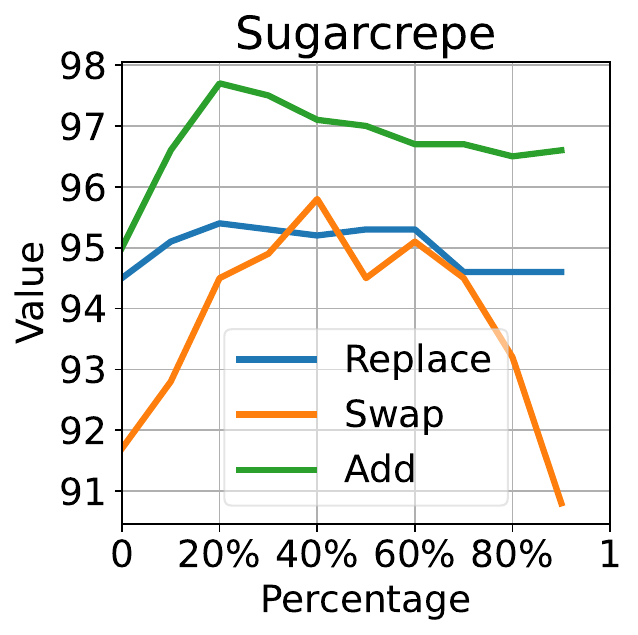}
    \includegraphics[height=8em]{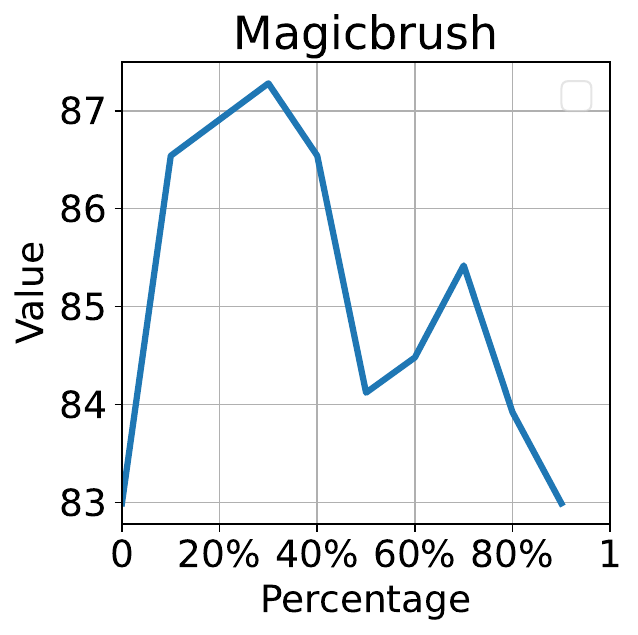}
    \vspace{-0.1in}
    \caption{Ablation on filter percentage of data. For most of the data, performance peaks around 30\% and reduces after that due to decrease in training data size.}
    \label{fig:ablation_filter}
    \vspace{-0.1in}
\end{figure}

\noindent\textbf{Effect of data filtering.} Fig.~\ref{fig:ablation_filter} shows the impact of removing the top $k$\% of biased data on the performance across four test datasets. Here, $k$ varies from 0 to 90, representing the progression from no data removal to the exclusion of 90\% of the data. The observed trend indicates that as the biased data is progressively removed, performance improves, peaking at approximately 30\% to 40\%. Beyond this point, performance declines due to the diminishing volume of training data.

We assess the quality of our filtered data by training a text-only model on 80\% of the filtered data and testing it on the remaining 20\%. Note that this evaluation is performed on filtered data, and a model should reach an accuracy close to 50\% in ideal data indicating it cannot distinguish between positive and negative caption just using text. We vary the parameter $k$ from 0\% to 90\%, in increments of 10\%, resulting in corresponding filtered data quality percentages of 75.9\%, 68.2\%, 60.7\%, 56.4\%, 53.8\%, 51.3\%, 50.6\%, 51.0\%, 49.8\%, and 47.9\%. Empirically, we find that data quality below 60\% is satisfactory for our purposes. Consequently, we decide to discard 30\% of the data in our experiment as it's a good trade-off between quality of filtered data and amount of training data.

To demonstrate that the distribution gap we identified also exists in datasets created by others, we apply the same evaluation method to  SugarCrepe~\cite{hsieh2023sugarcrepe}, a test benchmark also generated by GPT, that was refined using grammar and common sense models. Nevertheless, it still exhibits an accuracy of 69.0\%, indicating the presence of the bias. 

\noindent\textbf{Generalization on other model.} The bottom section of Table~\ref{table:ablation} shows that when BLIP2 is fine-tuned with the ITM head (a binary classification head) using our curated data, there is an observable improvement in performance. It's important to highlight that we only finetune and use the Q-former in BLIP2 without including any LLMs, and this model surpasses nearly all baselines in benchmarks that do not utilize LLMs (refer table~\ref{table:main}). As expected, the fine-tuned BLIP2 model does not outperform the model fine-tuned using LLaVA, particularly on the Winoground dataset, which is recognized for its challenging reasoning tasks. This suggests the importance of utilizing a more advanced vision-language model as the foundation model for achieving better results. Nonetheless, given that BLIP2's model size (180M) is significantly smaller compared to LLaVA-1.5 (13B), it offers a more lightweight option for simpler applications.

\begin{table}[t]
\caption{\textbf{Upper:} Ablating our training data based on LLaVA-1.5~\cite{liu2023improvedllava}. \textbf{Lower:} Finetuning a BLIP2 model with our data.}
\vspace{-5pt}
\centering
\resizebox{\columnwidth}{!}{
\begin{tabular}{@{}lccc|ccc|ccc|c@{}}
\toprule
{} & \multicolumn{3}{c|}{Winoground} & \multicolumn{3}{c|}{SeeTRUE} & \multicolumn{3}{c|}{SugarCrepe}  & {MagicBrush} \\

{} & {image} & {text} & {group} & DrawBench & EditBench & COCO-T2I & replace & swap & add  & {} \\
\midrule
{\color{gray}LLaVA-1.5~\cite{liu2023improvedllava}}           & {\color{gray}49.75} & {\color{gray}51.00} & {\color{gray}34.25} & {\color{gray}86.9}  & {\color{gray}78.3}  & {\color{gray}84.5}  & {\color{gray}93.5}  & {\color{gray}88.3}  & {\color{gray}95.8}  & {\color{gray}82.61} \\
Only replace w/ filter       & 51.25 & \textbf{54.25}  & 38.50  & 87.4  & 77.6  & 83.7  & 95.2  & 88.8  & 95.3  & 85.70 \\
Only swap w/ filter          & 63.50 & 49.25  & 42.25  & 81.0  & 74.0  & 79.2  & 92.6  & 95.5  & 91.3  & 76.63 \\
w/o filter                   & 65.75 & 51.50  & 46.75  & 88.4  & 76.4  & 81.1  & 94.5  & 91.7  & 95.0  & 82.99 \\
Random subsample   & 64.25 & 50.00  & 43.25  & 88.2  & 76.8  & 81.8  & 94.5  & 92.0  & 93.6  & 83.12 \\
\shortname{} (Ours)               & \textbf{68.00} & 53.75 & \textbf{47.25} & \textbf{88.8}  & \textbf{77.7}  & \textbf{84.9}  & \textbf{95.3}  & \textbf{94.9}  & \textbf{97.5}  & \textbf{87.28} \\
\midrule
BLIP2-ITM                               & 24.25 & 41.75  & 19.00  & 60.8  & 67.5  & 68.0  & 88.9  & 83.9  & 91.8  & 75.32 \\
BLIP2-ITM (ft on our data)              & \textbf{39.5} & \textbf{42.75} & \textbf{28} & \textbf{87.5}  & \textbf{76.2}  & \textbf{82.8}  & \textbf{94.3}  & \textbf{91.4}  & \textbf{96.0}  & \textbf{79.62} \\
\bottomrule

\end{tabular}

}
\vspace{1pt}
\label{table:ablation}
\end{table}

\subsection{Application: Text Alignment Ranking for Image Generation}

In this subsection, we showcase an application of our model for image generation. A major challenge with T2I models is their inconsistent adherence to complex compositional prompts~\cite{feng2023trainingfree}. Inspired by~\cite{karthik2023if}, one can generate a series of images from a T2I model and then employ our model to rerank them based on relevance, ultimately presenting the highest-ranked images to the user.

\begin{figure}[t!]
    \centering
    \includegraphics[width=0.85\textwidth]{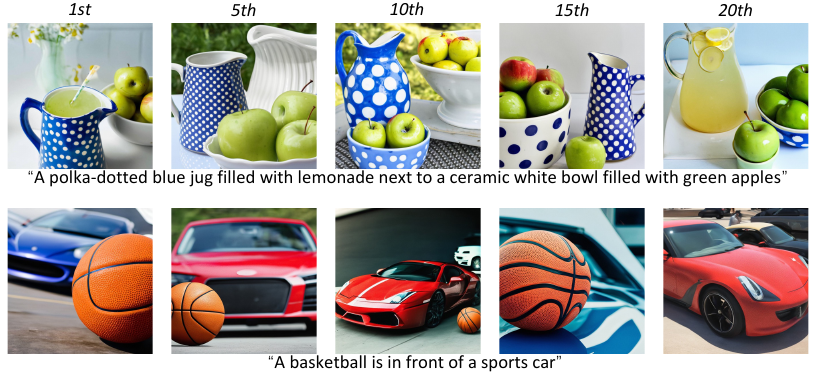}
    \caption{We show T2I generation results ranked by our model according to image-text alignment. The most aligned images are ranked higher. }
    \label{fig:ranking}
\end{figure}

TIFA~\cite{hu2023tifa} introduced a benchmark with images generated using different T2I models from identical prompts with human rankings. Our evaluation on this dataset reveals our model's superior alignment with human rankings, as shown in Table~\ref{table:tifa}. Spearman's $\rho$ and Kendall's $\tau$ are statistical measures that evaluate the similarity between two ranking orders, with higher values indicating greater alignment. Figure~\ref{fig:ranking} illustrates qualitative ranking results using our model for the generation results from a T2I model~\cite{if}.

\begin{table}[t]
\caption{Performance evaluation of ranking correlation on the TIFA~\cite{hu2023tifa} dataset.} 
\vspace{-5pt}
\centering
\resizebox{\columnwidth}{!}{
\setlength{\tabcolsep}{5pt}
\begin{tabular}{@{}lcccccccccc}
\toprule

{} & \makecell{CLIP-\\ViT-L-14~\cite{radford2021learning}} & \makecell{Neg\\CLIP~\cite{yuksekgonul2023when}} & \makecell{BLIP2-\\ITM~\cite{Li2023BLIP2BL}} & \makecell{BLIP2-\\ITC~\cite{Li2023BLIP2BL}} & \makecell{Image\\Reward~\cite{xu2024imagereward}} & \makecell{Visual\\GPT~\cite{lin2023visualgptscore}} & \makecell{VQ2\\(T5)~\cite{seetrue}} & \makecell{TIFA\\~\cite{hu2023tifa}} & \makecell{LLaVA-1.5\\~\cite{liu2023improvedllava}} & Ours\\
\midrule
Spearman $\rho$ & 28.5 & 40.1 & 43.3 & 51.5 & 62.8 & 36.9 & 55.1 & 59.7 & 61.5 & \textbf{64.5} \\
Kendall $\tau$  & 19.5 & 28.7 & 29.9 & 36.9 & 46.3 & 28.7 & 41.5 & 47.2 & 45.8 & \textbf{49.8} \\

\bottomrule

\end{tabular}

}
\vspace{1pt}
\label{table:tifa}
\end{table}

\section{Conclusion}
\vspace{-0.1in}
In this paper, we introduced an innovative method for generating high-quality training data for image-text alignment models. By implementing a mixed-type negative caption creation strategy and a novel filtering mechanism, we ensured a balanced distribution between positive and negative captions. This approach not only rectified the biases in existing models but also significantly enhanced the performance of our fine-tuned visual-language models, which achieve state-of-the-art results for image-text alignment. Our findings underscore the importance of high-quality, balanced training data in improving visual-language compositional reasoning and alignment. They also open up new avenues for exploring how these methods can be applied to other tasks or modalities in addition to the visual-language domain. In conclusion, our work offers an effective approach to improving image-text alignment, paving the road for future alignment studies.

This work was supported in part by NSF CAREER IIS2150012, Adobe Data Science award, Microsoft Accelerate Foundation Models Research Program, and Institute of Information \& communications Technology Planning \& Evaluation (IITP) grants funded by the Korea government (MSIT) (No. 2022-0-00871, Development of AI Autonomy and Knowledge Enhancement for AI Agent Collaboration) and (No. RS-2022-00187238, Development of Large Korean Language Model Technology for Efficient Pre-training).

\bibliographystyle{splncs04}
\bibliography{main}

\clearpage
\appendix
\section*{Appendix}

In this supplement, we will provide additional implementation details, present more results, and discuss limitations.

\section{More Implementation Details}

\textbf{Data filtering classifer} As stated in the main paper, to mitigate the bias arising from distribution discrepancies between positive and negative captions, which could influence the learning of vision-language models, we devised a strategy involving a text-only classifier. Our approach involves dividing the dataset into five equally sized partitions. In each iteration, we employ four partitions as the training set to fine-tune a DistilBERT model~\cite{Sanh2019DistilBERTAD}, with the fifth partition reserved as the test set. This cycle is repeated until every partition has been used as the test set. The fine-tuning parameters for the DistilBERT classifier are set to a learning rate of 2e-5 and a batch size of 16, over a duration of three epochs. Upon training, the classifier is applied to the hold-out test set to identify and filter out biased data, based on the classifier's prediction confidence. The data predicted correctly with the highest 30\% confidence will be excluded.

\textbf{BLIP2 finetuning} In Section 4.6 of the main paper, we further demonstrate the generalizability of our curated dataset by fine-tuning BLIP2~\cite{Li2023BLIP2BL} on it. Specifically, we fine-tuned BLIP2's Q-former using a batch size of 128 and a learning rate of 1e-5, across 1000 steps.

\textbf{Group Score for MagicBrush} Like the Winoground dataset, the MagicBrush consists of quartets that include one original image $I_{ori}$, one edited image $I_{edt}$, and their respective captions $C_{ori}$ and $C_{edt}$. We define text score and image score as:

\begin{equation}
f(C_{ori}, I_{ori}, C_{edt}, I_{edt}) = 
\begin{cases} 
1 & \text{if } s(C_{ori}, I_{ori}) > s(C_{edt}, I_{ori}) \\
0 & \text{otherwise}
\end{cases}
\end{equation}

and
\begin{equation}
g(C_{ori}, I_{ori}, C_{edt}, I_{edt}) = 
\begin{cases} 
1 & \text{if } s(C_{edt}, I_{edt}) > s(C_{edt}, I_{ori}) \\
0 & \text{otherwise}
\end{cases}
\end{equation}
respectively, where $s(C,I)$ represents the scoring function used to evaluate the alignment between a text-image pair. The group score is defined as follows: 

\begin{equation}
h(C_{ori}, I_{ori}, C_{edt}, I_{edt}) = 
\begin{cases} 
1 & \text{if } f(C_{ori}, I_{ori}, C_{edt}, I_{edt}) \text{ and } g(C_{ori}, I_{ori}, C_{edt}, I_{edt}) \\
0 & \text{otherwise} 
\end{cases}
\end{equation}

\begin{figure}[t!]
    \centering
    \includegraphics[width=0.95\textwidth]{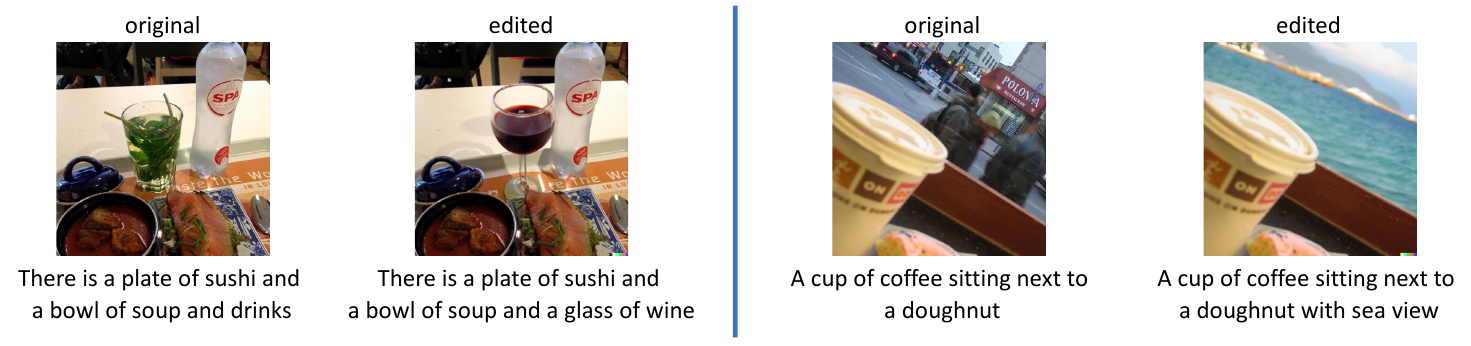}
    \vspace{-5pt}
    \caption{ \textbf{Magicbrush dataset} Often, the original caption remains consistent with the edited image.}
    \label{fig:magicbrush}
\end{figure}

Note that, contrary to the original definition from Winoground~\cite{thrush2022winoground}, we do not insist on $s(C_{edt},I_{edt}) > s(C_{ori},I_{edt})$ for the test score or $s(C_{ori},I_{ori}) > s(C_{ori},I_{edt})$ for the image score. This adjustment is made because, within the Magicbrush dataset, it is often observed that the original image's caption $C_{ori}$ aligns well with the edited image $I_{edt}$, as demonstrated in Figure~\ref{fig:magicbrush}. Consequently, we do not penalize a model for assigning a high score to $s(C_{ori},I_{edt})$.

\begin{table}[t!]
\caption{Comparison of our model against baselines across three prevalent challenges, highlighting the highest image score in bold.}
\centering
\scalebox{0.9}{
\setlength{\tabcolsep}{5pt}
\begin{tabular}{@{}lc|c|c@{}}
\toprule
{} & {Attribute} & {Counting} & {Spatial} \\
\midrule
CLIP-ViT-L-14~\cite{radford2021learning}           & 70 & 60 & 56 \\
NegCLIP~\cite{yuksekgonul2023when}                 & 76 & 62 & 62 \\
BLIP2-ITC~\cite{Li2023BLIP2BL}               & 80 & 60 & 64 \\
VisualGPT~\cite{Chen2022CVPR}              & \textbf{98} & 82 & 78 \\
BLIP2-ITM~\cite{Li2023BLIP2BL}           & 90 & 82 & 58 \\
Image-Reward~\cite{xu2024imagereward}        & 94 & 76 & 58 \\
VQ2 (BLIP-T5)~\cite{seetrue}       & 90 & 78 & 60 \\
LLaVA-1.5~\cite{liu2023improvedllava}           & 90 & 80 & 80 \\
Ours                & \textbf{98} & \textbf{84} & \textbf{90} \\
\bottomrule
\end{tabular}
}
\vspace{-0.1in}
\label{table:mydata}
\end{table}

\section{More Results}

\textbf{Image Score for Attribute, Counting and Spatial Reasoning} In Section 4.5 of the main paper, we specifically focus on the attribute, counting, and spatial reasoning capabilities of vision-language models by curating a custom dataset. This dataset is generated by first creating captions using GPT, which are then used to synthesize images via a Text-to-Image (T2I) model~\cite{if}. For each generated image pair, we manually select one image that aligns well with the caption (positive) and one that does not (negative). While the main paper presents our results in terms of accuracy, in this section, we introduce an image score metric. For each text-images triplet, we assign a score of 1 if $s(C,I_{pos}) > s(C,I_{neg})$, and 0 otherwise. Table~\ref{table:mydata} shows our results. Again, our model achieves the best performance across all datasets.

\textbf{Contrastive Training} The results we have discussed so far are based on models trained with the original cross-entropy loss. Inspired by the CLIP~\cite{radford2021learning}, we have also attempted to fine-tune the LLaVA-1.5 model using our data with a contrastive loss approach. However, we empirically found that the contrastive loss was not as effective as anticipated: (67, 54.24, 46.5) (88.9, 77.3, 84.2)  (95.5, 94.8, 97.6) (86.98) (compare with last row in Table1 in main paper). We hypothesize that the current cross-entropy loss might produce a similar effect to contrastive loss when dealing with two types of data (positive vs negative), and contrastive loss might be more beneficial when there is a spectrum of ``negativeness," e.g., where some pairs are completely unaligned and others are only slightly incorrect. To explore this, we created three types of data: positive (aligned), slightly unaligned (the curated data in our paper), and completely unaligned (randomly shuffled image-text pairs), aiming for progressively lower scores across these categories. However, this approach proved unhelpful, as the last case was too straightforward for LLaVA to discern, providing no training signal. Thus, exploring how to create a spectrum of data alignment levels and studying if they are useful remains an interesting future research.

\begin{figure}[t!]
    \centering
    \includegraphics[width=0.95\textwidth]{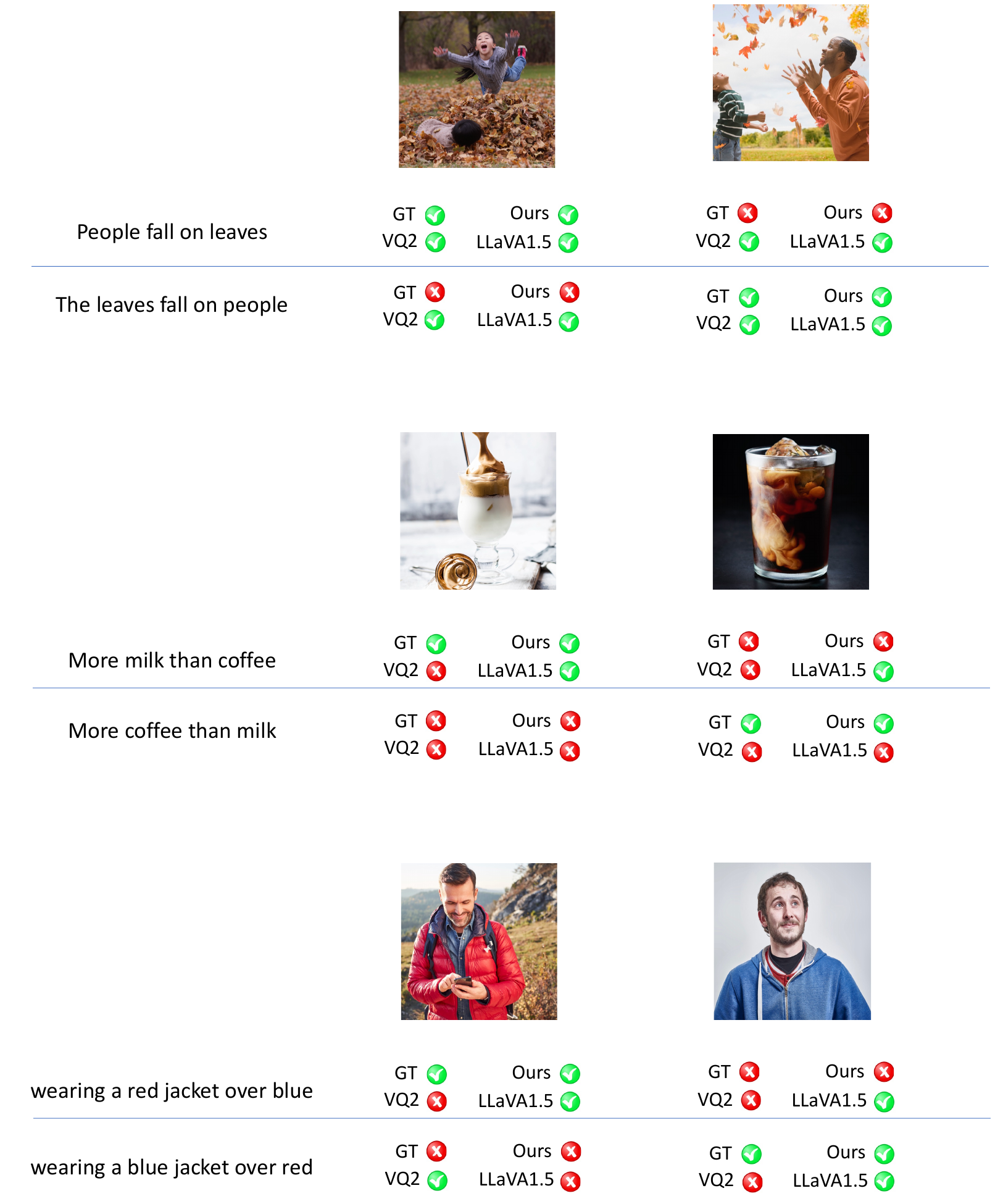}
    \vspace{-5pt}
    \caption{Qualitative Results on Winoground}
    \label{fig:winogroud_result}
\end{figure}

\textbf{Qualitative Results on Winoground} Figure~\ref{fig:winogroud_result} presents qualitative results for the Winoground dataset, comparing our model against the strongest baseline, LLaVA-1.5~\cite{liu2023improvedllava}, and the recently introduced VQ2~\cite{seetrue}. The outcomes demonstrate that our model exhibits superior compositional understanding abilities.

\section{Discussion}

In this paper, we introduce a novel approach for generating high-quality training data for image-text alignment models. However, our method has some limitations. Firstly, it assumes access to ground truth positive captions for images, which may not be feasible on a large scale. A promising avenue for overcoming this hurdle could involve leveraging vision-language models such as LLaVA for generating image captions, presenting an exciting direction for future research to enhance scalability. Additionally, our method may inherit certain constraints from the LLaVA image encoder, which utilizes CLIP and is tailored to process images of a fixed, relatively low resolution. Consequently, this may limit the model's ability to detect small objects or capture fine-grained details effectively.

\end{document}